\documentclass[runningheads]{llncs}
\usepackage{booktabs}
\usepackage{cite} % get in-order ref index
\usepackage{enumitem}
\setitemize{label=\textbullet} % get point \item

\usepackage{multirow}
\usepackage[T1]{fontenc}
\usepackage{graphicx}
\begin{document}
\title{GLoRE: Evaluating Logical Reasoning of Large Language Models}

\author{Hanmeng Liu\inst{1}\and
Zhiyang Teng\inst{3}
Ruoxi Ning\inst{2} \and
Yiran Ding\inst{2} \and
Xiulai Li\inst{1} \and
Xiaozhang Liu\inst{1} \and
Yue Zhang\inst{2}
}
\authorrunning{H. Liu et al.}
\institute{Hainan University, Haikou, Hainan \and Westlake University, Hangzhou, Zhejiang, China \and ByteDance SG}
\maketitle              % typeset the header of the contribution
\begin{abstract}
Large language models (LLMs) have shown significant general language understanding abilities. However, there has been a scarcity of attempts to assess the logical reasoning capacities of these LLMs, an essential facet of natural language understanding. 
To encourage further investigation in this area, we introduce GLoRE, a \textbf{G}eneral \textbf{Lo}gical \textbf{R}easoning \textbf{E}valuation platform that not only consolidates diverse datasets but also standardizes them into a unified format suitable for evaluating large language models across zero-shot and few-shot scenarios.
Our experimental results show that compared to the performance of humans and supervised fine-tuning models, the logical reasoning capabilities of large reasoning models, such as OpenAI's o1 mini, DeepSeek R1 and QwQ-32B, have seen remarkable improvements, with QwQ-32B achieving the highest benchmark performance to date.
%In addition, to understand the effect of in-domain training, we fine-tuned an open 7b-sized LLM and observed a performance gain.
GLoRE is designed as a living project that continuously integrates new datasets and models, facilitating robust and comparative assessments of model performance in both commercial and Huggingface communities. It garnered over 300 citations upon its release.

\end{abstract}
\section{Introduction}

Large Language Models (LLMs)~\cite{openai2023gpt4,touvron2023llama}, especially reasoning language models~\cite{deepseekai2025deepseekr1incentivizingreasoningcapability,openai2024openaio1card} demonstrate advanced capabilities in complex reasoning tasks and show significant adaptability and versatility across various applications, from simple everyday tasks to specialized domains such as coding, mathematics, law, medicine, and finance~\cite{choi2023chatgpt,frieder2023mathematical,kung2023performance,doi:10.1126/science.abq1158,wu2023bloomberggpt}.
Quantitative evaluation of LLM reasoning has thus become a very important task. To this end, existing work has considered mathematical reasoning~\cite{cobbe2021training,hendrycks2021measuring}, algorithmic problem solving~\cite{chen2021codex,quan2025codeelo}, and knowledge-driven reasoning~\cite{hendryckstest2021,wang2024mmlu}.

\begin{figure*}[t!]
\centering 
\setlength{\abovecaptionskip}{0.1cm}
\setlength{\belowcaptionskip}{0.2cm}
\includegraphics[width=0.7\textwidth]{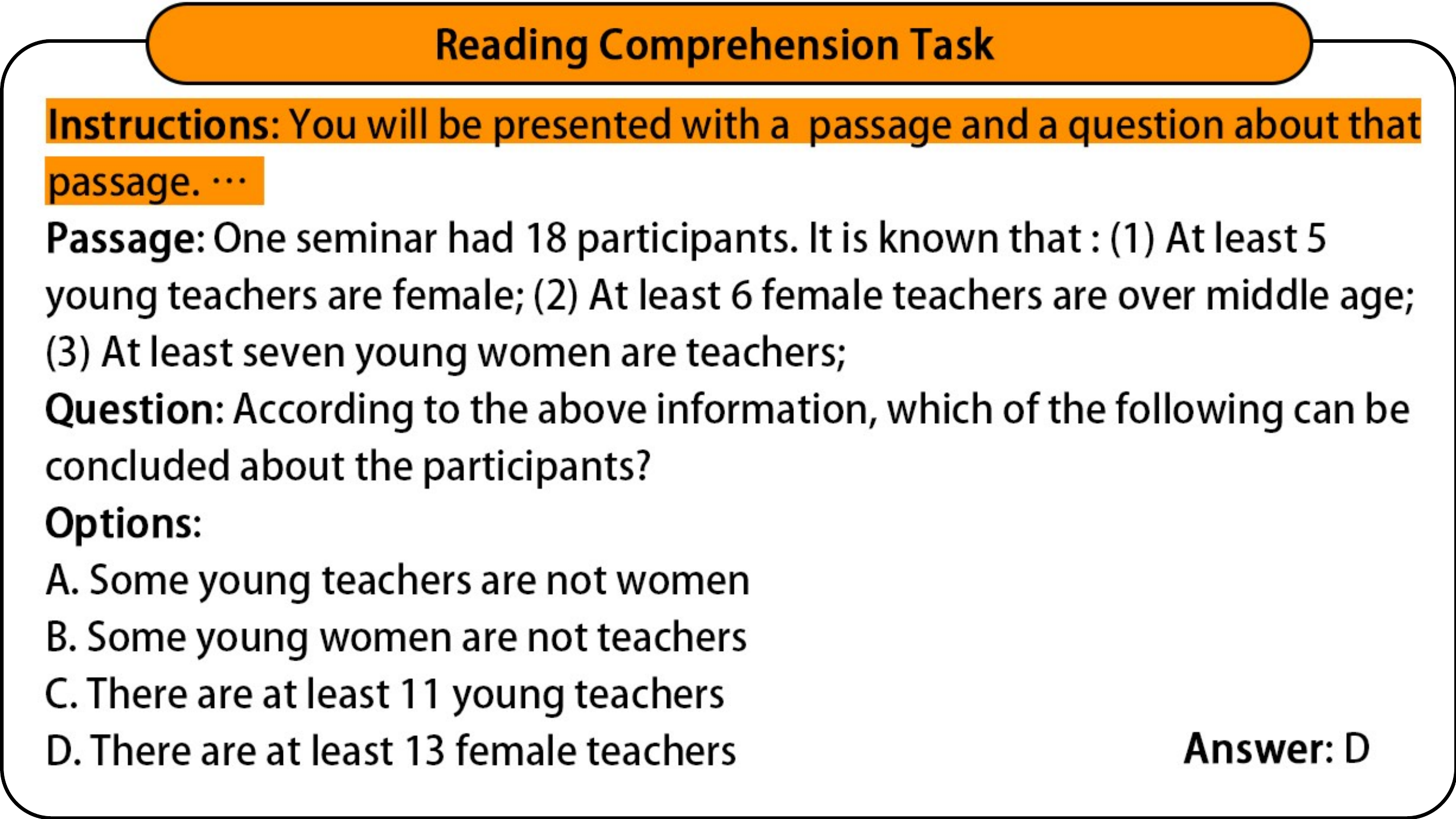}
\caption{Instruction and question format for logical reading comprehension tasks.}
\label{fig:showcase}
\vspace{-2em}
\end{figure*}

Logical reasoning is a cornerstone of human intelligence and has been a central focus in artificial intelligence research since its inception~\cite{cresswell1973logics,iwanska1993logical,kowalski1979logic}. 
However, evaluating verbal reasoning turned out to be too difficult in the 1950s due to insufficient natural language understanding (NLU) technologies, and thus AI researchers focused on formal logical reasoning instead~\cite{iwanska1993logical,McCHay69,1056797}. Since the 2010s, NLU has witnessed huge advances, where reading comprehension~\cite{sar,drop} and natural language inference~\cite{bowman-etal-2015-large,N18-1101} tasks were solved with high accuracies, which made verbal reasoning evaluation feasible~\cite{yu2020reclor,DBLP:journals/corr/abs-2007-08124}.
Figure~\ref{fig:showcase} illustrates an example of logical reasoning in reading comprehension. To address such questions, LLMs must engage in multi-step, algorithmic, symbolic reasoning. This makes logical reasoning an ideal testbed for evaluating LLMs' ability to process complex natural language information accurately, robustly, and logically.

To this end, we introduce the General Logical Reasoning Evaluation (GLoRE) benchmark, designed to assess instruction-tuned LLMs on various logical reasoning tasks. GLoRE evaluates the strengths and limitations of LLMs in this domain, similar to how GLUE~\cite{wang-etal-2018-glue} and Super-GLUE~\cite{10.5555/3454287.3454581} benchmark natural language understanding. GLoRE includes three types of logical reasoning tasks: {Multi-choice Reading Comprehension}~\cite{race}, {Natural Language Inference (NLI)}~\cite{Dagan2005ThePR}, and {True-or-False (Yes-or-No) Questions}~\cite{clark2019boolq}. These tasks encompass a wide range of logical reasoning phenomena, with high-quality datasets that remain challenging for pre-trained language models \cite{huang2023reasoning,clark2019boolq,koreeda-manning-2021-contractnli-dataset}. In total, GLoRE covers 12 datasets with 72,848 instances.
Since its release in 2023, GLoRE has been used for evaluating language models, receiving over 300 citations on ArXiv.
% 第二页第二段最后一句话，应该把下载量写出来，入选了哪些大榜单写出来，然后引用的数量不要写“大约”。[ ]

% 第二页第三段第四行开头有个标点符号不对。[v]
Using GLoRE, we report the logical reasoning capabilities of commercial models such as GPT-4 and OpenAI o1~\cite{openai2024openaio1card}, as well as popular open-source models such as LLaMA \cite{touvron2023llama}, Falcon \cite{falcon40b}, Mistral \cite{jiang2024mixtral}, DeepSeek R1~\cite{deepseekai2025deepseekr1incentivizingreasoningcapability}, and QwQ-32B~\cite{qwq32b}. We test their instruction-following and problem-solving abilities in logical reasoning tasks. Results show that while commercial LLMs like GPT-4 still excel in zero-shot settings and approach human performance on specific datasets like ReClor, open-source models like QwQ-32B now rival or even surpass commercial counterparts in key tasks, achieving state-of-the-art results on multiple benchmarks. This underscores rapid advancements in open-source LLMs, narrowing the performance gap with commercial models.
% 第二页第三段第六行是不是应该把那些推理模型加进来。[v]
However, performance varies significantly across datasets, indicating sensitivity to data distribution. This sensitivity is further confirmed by observations that in-context learning and supervised fine-tuning primarily enhance LLM performance on specific test distributions, demonstrating their strong learning ability. While LLMs show promise in logical reasoning, their robustness to data distribution variations remains a challenge, highlighting the need for further improvement.

%To our knowledge, GLoRE is the first instruction-prompt evaluation suite for logical reasoning, and we are the first to evaluate LLMs' complex logical reasoning abilities comprehensively. We release our benchmark at \url{https://anonymous.com}.

% \vspace{-4mm}
\section{Related Work}

\textbf{Logical reasoning with natural language.} Tapping into logical reasoning capabilities represents a holistic endeavour in natural language understanding (NLU). A variety of methods have been explored to realize this objective, including symbolic systems~\cite{article,Poole1987,maccartney-manning-2007-natural}, fine-tuning of language models~\cite{wang-etal-2018-glue,huang2021dagn,Xu_2022,liu2023logicot}, and hybrid approaches combining neural and symbolic elements~\cite{li-srikumar-2019-augmenting,saha2020prover,sanyal2022fairr}. 

%Traditionally, in the realm of logic and semantics, computational linguists crafted symbolic systems grounded in First-Order-Logic (FOL) or Natural Logic~\cite{maccartney-manning-2007-natural} to address essential inference tasks. However, these rule-based models found it challenging to tackle problems such as the Recognizing Textual Entailment (RTE) challenge~\cite{Dagan} using hand-crafted rules and theorem provers alone.

The recent introduction of evaluation datasets, notably LogiQA~\cite{DBLP:journals/corr/abs-2007-08124} and Reclor~\cite{yu2020reclor}, has reinvigorated the focus on logical reasoning in NLP research. Logical reasoning is now leveraged in numerous probing tasks over large Pre-trained Language Models (PLMs) and applied to downstream tasks such as question-answering and dialogue systems \cite{shi-etal-2021-neural,beygi2022logical}. Despite these advancements, the aspiration to emulate human-like logical reasoning capabilities within NLU systems remains a significant challenge for traditional models~\cite{DBLP:journals/corr/abs-2007-08124,huang2023reasoning}. In this study, our goal is not only to quantitatively evaluate the capability of Large Language Models (LLMs) in addressing the previously mentioned challenge but also to underscore the significance of our work in providing a validated platform for enhancing various reasoning methods with our data.

\textbf{LLM reasoning evaluation.} 
Despite progress in evaluating LLMs for specific reasoning tasks like arithmetic \cite{qin2023chatgpt} and commonsense \cite{bang2023multitask}, a yawning gap exists in comprehensively assessing their logical reasoning. While LLMs excel at specific tasks like arithmetic reasoning \cite{qin2023chatgpt}, they face challenges in complex areas like multi-step reasoning \cite{fu2023chain} and abstract scenarios \cite{gendron2023large}. ChatGPT exhibits strengths in chat-specific reasoning and some commonsense domains \cite{bang2023multitask,ott2023thoughtsource}, but struggles with tasks requiring longer chains of inference \cite{bang2023multitask}. Other LLMs like FLAN-T5~\cite{chung2022scaling}, LLaMA~\cite{touvron2023llama}, and PaLM~\cite{anil2023palm} show potential in general deductive reasoning \cite{saparov2023testing}, while InstructGPT and Codex excel in specialized domains like medical reasoning \cite{lievin2022can}. Despite these advances, limitations in data bias \cite{orru2023human}, and complex reasoning tasks necessitate further research and optimization to fully unlock the reasoning potential of LLMs \cite{wu2023reasoning}.

Big-Bench Hard (BBH)~\cite{suzgun2022challenging} isolates 23 most challenging tasks from BIG-Bench~\cite{srivastava2023beyond}. These tasks comprise general language understanding, arithmetic and algorithmic reasoning, and logical deduction. However, in comparison to our benchmark, the data size of the logical reasoning section in BBH is very small.
HumanEval~\cite{chen2021evaluating} serves as a hand-written evaluation set for coding. The programming problems included are designed to assess language comprehension, reasoning, algorithms, and simple mathematics. While similar to logical reasoning in that code generation necessitates complex reasoning skills, GLoRE differs in presenting logical reasoning problems via natural language prompts.

\begin{table*}
\small
\centering
\scalebox{0.95}{
\begin{tabular}{lcc || lcc}  % @{\hspace{2em}} adds horizontal separation
\hline
\textbf{Dataset} & \textbf{Size} & \textbf{Target} & \textbf{Dataset} & \textbf{Size} & \textbf{Target} \\
\hline
LogiQA 2.0 test & 1,572 & 4-way multi-choice & ConTRoL & 805 & E, C, N \\
LogiQA 2.0 zh test & 1,594 & 4-way multi-choice & HELP & 35,891 & E, C, N \\
ReClor dev & 500 & 4-way multi-choice & TaxiNLI test & 10,071 & E, C, N \\
AR-LSAT test & 230 & 5-way multi-choice & NaN-NLI & 259 & E, C, N \\
LogiQA22 & 1,354 & 4-way multi-choice & FraCas & 346 & Yes, No, Neutral \\
&&&RuleTaker dev & 10,068 & Yes, No \\
&&&ProofWriter dev & 10,158 & Yes, No \\
\hline
\end{tabular}
}
\vspace{0.2em}
\caption{\label{Tab:stat}
Data statistics. (``E'': entailment; ``C'': contradiction; ``N'': neutral.)
}
\vspace{-2em}
\end{table*}

ARB~\cite{sawada2023arb} is a benchmark for advanced reasoning over multiple fields like mathematics, physics, biology, chemistry, and law. Similar to GLoRE, it introduces a challenging subset of math and physics problems that require advanced symbolic reasoning. However, the benchmark constraints its problem on the above subjects with domain knowledge, not general logical reasoning questions, which is the focus of GLoRE.

\section{The GLoRE Dataset}
\label{sec:data}

%Similar to GLUE~\cite{wang-etal-2018-glue} and~SuperGLUE \cite{10.5555/3454287.3454581}, GLoRe functions as an assembled dataset for specialized logic reasoning  evaluations. 
As mentioned in the introduction, GLoRE contains three NLU tasks: Multi-choice Reading Comprehension, NLI, and Yes-or-No. First, Multi-choice reading comprehension ~\cite{race} is essential in verbal reasoning tests, which cover abundant high-quality logical reasoning problems in the wild. 
Second, Unlike multi-choice reading comprehension, NLI \cite{Dagan2005ThePR} is more general and centric on entailment relations in a simpler task format, which is a fundamental task for evaluating reasoning abilities \cite{poliak-etal-2018-collecting,demszky2018transforming}.
Third, the Yes-or-No reasoning task \cite{clark2019boolq} is a combination of question-answering and textual entailment, which can serve as a playground for testing models’ reasoning abilities~\cite{ijcai2020-537,tafjord2021proofwriter}. 
The data statistics are shown in Table~\ref{Tab:stat}. 
% 第四页表格和文字之间的空白太小了。[v]
% We delve into its components in the subsequent sections, detailing the Multi-choice Reading Comprehension (Section~\ref{sec:mrc}), Natural Language Inference (Section~\ref{sec:nli}), and True-or-False (Section~\ref{sec:torf}) test sets. Finally, we discuss our prompt design for the evaluation of LLMs (Section~\ref{sec:prompt}).

\subsection{Multi-choice Reading Comprehension (MRC)}
\label{sec:mrc}
Within the standard multi-choice reading comprehension (MRC) task setting, a system is presented with a passage and a question, and the objective is to choose the most suitable answer from a set of candidate responses. 
% An example of logical MRC can be seen in Figure~\ref{fig:showcase}. 
% 3.1节第三行“图一展示了阅读理解的案例”这句话应该可以删掉。[v]
Particularly, GLoRE contains five such datasets:
% 我觉得这些列表应该统一，应该用列表表格式，每段前边有个小圆点来列举。现在不同的数据评价基准前面缩进都不一样。[v]
% \begin{itemize}
%\item 

\noindent \textbf{LogiQA}~\cite{DBLP:journals/corr/abs-2007-08124} is a logical MRC dataset derived from the Chinese Civil Service Examination, translated into English, and made available in both Chinese and English versions. 
We adopt LogiQA 2.0 \cite{10174688} and use both the English (\textbf{LogiQA 2.0}) and  Chinese (\textbf{LogiQA 2.0 zh}) test sets for our evaluation. %As shown in Table~\ref{Tab:stat}, LogiQA 2.0 test contains 1,572 instances and LogiQA 2.0 zh test contains 1,594 instances.
%\footnote{\url{https://github.com/csitfun/LogiQA2.0}}
% 3.1节第二段the second version直接改成2.0就能节省一行。[v]

% \item 
\noindent \textbf{ReClor}~\cite{yu2020reclor} comprises question-answering examples from the LSAT exams designed to assess human logical reasoning abilities. We use the development set for our testing as the test set does not provide gold labels. %As shown in Table~\ref{Tab:stat}, ReClor dev has 500 instances.
%\footnote{\url{https://www.lsac.org/lsat}}

% \noindent 
\noindent \textbf{AR-LSAT}~\cite{wang2022lsat} is a dataset of analytical reasoning questions from the Law School Admission Test. 
%It encompasses 2,064 questions, each outlining a reasoning game that falls into one of three primary types: (1) ordering game, (2) grouping game, and (3) assignment game. 
Each question contains five options rather than four.  

\iffalse 
\noindent \textbf{LogiQA22} is a dataset collected and processed according to the LogiQA format. It incorporates the newly released Chinese Civil Servant Exams from 2022, which are not included in the original LogiQA 2.0 dataset.\footnote{As of the current date, ChatGPT and GPT-4 are limited to training data up until September 2021.(\url{https://help.openai.com/en/articles/6783457-what-is-chatgpt}).} As a result, it can serve as an test set for ChatGPT and GPT-4. To illustrate the difference between the LogiQA22 data and the LogiQA data, we annotate the reasoning types of the LogiQA22 data following \citet{DBLP:journals/corr/abs-2007-08124}. Figure~\ref{fig:distribution} in Appendix~\ref{appendix_type} compares the logic reasoning type distribution of LogiQA 2.0 and LogiQA22. The number of training examples of LogiQA22 is similar to that of the LogiQA 2.0 test set (1,354 vs. 1,572, as shown in Table~\ref{Tab:stat}). However, LogiQA22 contains significantly less categorical reasoning and more sufficient conditional reasoning, necessary conditional reasoning, and disjunctive reasoning examples than the LogiQA 2.0 test set.
% making it suitable for accessing OOD performances. 
\input{Figures/distribution}
\fi 

% \begin{figure}[t!]
% \centering
% \setlength{\abovecaptionskip}{0.1cm}
% \setlength{\belowcaptionskip}{0.2cm}
% \includegraphics[width=0.4\textwidth]{Figures/distribution_bar.pdf}
% \caption{Percentages of reasoning types for the LogiQA 2.0 test and the LogiQA22 data.  }
% \label{fig:distribution}
% \vspace{-2em}
% \end{figure} 

% \noindent 
\noindent \textbf{LogiQA22} is collected and processed according to the LogiQA 2.0 format after ChatGPT was released. It incorporates the newly released Chinese Civil Servant Exams from 2022, which are not included in the original LogiQA dataset.
%\footnote{As of the current date, ChatGPT and GPT-4 are limited to training data up until September 2021. (\url{https://help.openai.com/en/articles/6783457-what-is-chatgpt}).}  
%We annotate the reasoning types of the OOD data following \citet{DBLP:journals/corr/abs-2007-08124}. 
% Figure~\ref{fig:distribution} compares the logic reasoning type distribution of the LogiQA 2.0 test set and LogiQA22, which shows one aspect of distributional variation between the two datasets. As we can see from the figure, LogiQA22 contains significantly less categorical reasoning and more sufficient conditional reasoning, necessary conditional reasoning, and disjunctive reasoning examples than the LogiQA test set, making it suitable for assessing ChatGPT performance to unseen data. 
% %As a result, it can be served as an out-of-distribution (OOD) test set for GPT-4. 
% The number of test examples of LogiQA22 is similar to that of the LogiQA 2.0 test set (1,354 vs. 1,572 as shown in Table~\ref{Tab:stat}).
% \end{itemize}

%\input{Figures/distribution}
%% The test set is a collection of logical reasoning tests designed by experts from 2022 onwards. We release our data at \url{https://github.com/csitfun/LogiQA2.0}.

% \begin{figure}[htbp]
% \centering
% \setlength{\abovecaptionskip}{0.1cm}
% \setlength{\belowcaptionskip}{0.2cm}
% % \includegraphics[width=0.48\textwidth]{control.png}
% \includegraphics[width=0.48\textwidth]{Figures/control.pdf}
% \caption{An NLI example from the ConTRoL dataset.}
% \label{fig:control}
% %\vspace{-0.5cm}
% \end{figure} 

\subsection{Natural Language Inference (NLI)}
\label{sec:nli}

NLI is the task of determining the logical relationship between a hypothesis and a premise. The typical scheme involves text classification, where the model selects one of three labels: \emph{entailment}, \emph{contradiction}, and \emph{neutral}. 
% \begin{itemize}

% \noindent 
\noindent \textbf{ConTRoL} \cite{DBLP:journals/corr/abs-2011-04864} is an NLI dataset that offers an in-depth examination of contextual reasoning within the NLI framework. 
Approximately 36.2\% of premise-hypothesis pairs fall under the category of logical reasoning in this dataset.
We choose the logical reasoning portion for our evaluation. %As shown in Table~\ref{Tab:stat}, it has 805 instances.

% \noindent 
\noindent \textbf{HELP}~\cite{yanaka-EtAl:2019:starsem} is an NLI dataset emphasizing monotonicity reasoning, a crucial concept in Natural Logic \cite{maccartney2007natural}. 
We use the training set for our evaluation. 
%The datasets are generated through monotonicity rules and specifically scrutinize inferences related to monotonicity.
%As shown in Table~\ref{Tab:stat}, it has 35,891 instances.
%\textbf{ConjNLI} \cite{DBLP:journals/corr/abs-2010-10418} is a challenging stress test for NLI over conjunctive sentences, where the premise differs from the hypothesis by having conjuncts being removed, added, or replaced. Logical reasoning about conjunctions is heavily tested in ConjNLI. Premise-hypothesis pairs are created automatically by applying conjunct operations on collected conjunctive sentences. 
%Here is an example from the ConjNLI dataset:

%Premise: In Quebec, an allophone is a resident, usually an immigrant, whose mother tongue or home language is neither French nor English.

%Hypothesis: In Quebec, an allophone is a resident, usually an immigrant, whose mother tongue or home language is not French.

%Label: Entailment

% \noindent 
\noindent \textbf{TaxiNLI}~\cite{DBLP:journals/corr/abs-2009-14505} is an NLI dataset that has been re-annotated based on MNLI~\cite{williams-etal-2018-broad}, with categories include logical categories such as connectives, mathematical reasoning, and deduction. 
%These annotations include logical categories such as connectives, mathematical reasoning, and deduction. 
%As shown in Table~\ref{Tab:stat}, TaxiNLI test has 230 instances.

% \noindent 
\noindent \textbf{NaN-NLI}~\cite{truong-etal-2022-another} is a test suite designed to probe the capabilities of NLP models in capturing sub-clausal negation. 
%Understanding and interpreting sub-clausal negation in a text involves logical reasoning, as it requires the ability to identify and handle the logical operator of negation within the context of a sentence or a clause. Hence, 
The successful handling of sub-clausal negation can be seen as a strong indicator of a model's logical reasoning capacity. %As shown in Table~\ref{Tab:stat}, NAN-NLI has 259 instances.
% \end{itemize}
% \begin{figure}[htbp]
% \centering
% \setlength{\abovecaptionskip}{0.1cm}
% \setlength{\belowcaptionskip}{0.2cm}
% \includegraphics[width=0.48\textwidth]{Figures/fracas.pdf}
% \caption{An example from the FraCaS dataset.  }
% \label{fig:fracas}
% %\vspace{-0.5cm}
% \end{figure} 
\subsection{True-or-False (Yes-or-No) Questions (TF)}
\label{sec:torf}
% \begin{itemize}
% \noindent 
% 3.3节第一个自然段也可以想办法缩减一行。[ ]
\noindent \textbf{FraCaS} test suite~\cite{Pulman1996UsingTF} presents complex entailment problems involving multi-premised contexts as a three-way classification task. The ability to determine entailment relationships in this context is closely tied to logical reasoning. 
%We convert the ``Don't know'' label into a single ``Neutral'' token. %As shown in Table~\ref{Tab:stat}, FraCas has 346 instances.

% \noindent 
\noindent \textbf{RuleTaker}~\cite{ijcai2020-537} dataset is a synthetic creation designed to examine the reasoning ability of transformer models~\cite{DBLP:journals/corr/VaswaniSPUJGKP17} over natural language rules.
%It provides input facts and rules as context, requiring a binary true-or-false response as output. 
%Despite its original design for question-answering, the dataset can be readily converted into an NLI-style binary classification task. 
This task explicitly targets logical reasoning by asking models to reason over a set of rules and facts to generate true-or-false responses as output. 

% \noindent 
\noindent \textbf{ProofWriter}~\cite{tafjord2021proofwriter} dataset generates sets of facts and rules, each followed by questions, which can be proven true or false using proofs of various depths. 
%This directly targets logical reasoning by asking models to generate proofs to validate or invalidate a given statement. We choose the two-way classification version of the dataset for our evaluation. 
% \end{itemize}
%\input{subtexs/4-method}

%\input{subtexs/method}

\section{Experiments}
% 第四节第一个自然段应该增加一些引用。[ ]
% We use GLoRE to evaluate logical reasoning of traditional pre-trained models, and reasoning enhanced LLMs, including open-sourced and closed LLMs, comparing the results with human performance.
We employ GLoRE to assess the logical reasoning capabilities across different categories of language models, including traditional pre-trained models and reasoning-enhanced LLMs, both open-source and proprietary. We conduct a comprehensive comparative analysis of their performance against human benchmarks.
\subsection{Experimental Settings}
We adopted \textbf{RoBERTa-base}~\cite{liu2019roberta} as a baseline, fine-tuning it on the training set over five epochs for each dataset. 
The community models selected for comparison include \textsc{Falcon-40b-instruct}~\cite{falcon40b}, \textsc{LLaMA-30b-supercot}~\cite{touvron2023llama} \textsc{Mixtral-8x7b}, \textsc{DeepSeek R1}~\cite{deepseekai2025deepseekr1incentivizingreasoningcapability} and QwQ-32B~\cite{qwq32b}. 
For OpenAI models, we choose \textsc{ChatGPT}, \textsc{GPT-4} and \textsc{o1 mini}\cite{openai2024openaio1card}.
%Both \textbf{ChatGPT} and \textbf{GPT-4} are evaluated with the OpenAI Evaluation framework\footnote{\url{https://github.com/openai/evals}}, a comprehensive tool designed for the evaluation of OpenAI models. The specific versions of the models assessed are labeled as "gpt-3.5-turbo-0301" for ChatGPT and "gpt-4-0314" for GPT-4, respectively.
%\footnote{We had been granted early access to the GPT-4 API which enabled us to utilize it within the OpenAI Evaluation framework.} 
%Moreover, we engage the GPT-4 Chat UI to conduct a series of case studies on GPT-4. These examinations probe into the model's in-context learning abilities and chain-of-thought reasoning capabilities, by using two OpenAI Plus accounts.%\footnote{This research was conducted in Singapore in alignment with OpenAI's policy.}

%All experiments were executed on 40G VRAM A100 GPUs based on the HuggingFace transformers library. 
Our evaluation metrics consisted of classification accuracy scores. Additionally, we utilized reported accuracies for datasets where human performance data was available and recorded both the average and peak performance of human participants to establish a human baseline. For the LogiQA22 dataset, we engaged five co-authors as test subjects and computed their accuracy based on 150 test examples. 
%500 RMB (about 20 test examples per hour) is spent with an average pay rate of 67 RMB (about 10 USD) per hour, which is above the domestic hourly wage.

%\noindent \textbf{Instruction and Prompt Design} Instruction-tuned LLMs are specifically trained to follow natural language instructions. In this setup, the task input is transposed into a prompt via templates, and the gold label is verbalized~\cite{liu2021pretrain}, and the methods of prompting have been found to significantly influence model performance. Prior research indicated that ChatGPT could underperform in question-answering scenarios if the instruction was not appropriately optimized~\cite{zhong2023agieval}. Consequently, we conducted an investigation into different zero-shot prompting methods to enhance the performance of the models being tested. We specifically explored prompting design methods for Natural Language Inference (NLI) tasks. \citet{qin2023chatgpt} employed a binary scheme that labeled the model outputs as either "entail" or "not entail". Since the NLI datasets in GLoRE encompass three-way classification tasks, we therefore prompted models to articulate the following three potential relations between the premise and hypothesis: entailment, contradiction, and neutrality.
%The finalized instructions for the three types of tasks are presented in Appendix \ref{sec:appendix_b}.

\begin{table*}[!t]
\footnotesize
\scalebox{0.8}{
\begin{tabular}{@{}l|ccccc|cccc|ccc|c}
\toprule
\textbf{Task}&\multicolumn{5}{|c|}{\textbf{MRC}}&\multicolumn{4}{|c|}{\textbf{NLI}}&\multicolumn{3}{|c|}{\textbf{TF}}& \multirow{2}{*}{\textbf{Avg}} \\

\cmidrule(r){1-13}
\textbf{Dataset} & \textbf{LQ} & \textbf{LQ zh}& \textbf{RC} & \textbf{AL} & \textbf{LQ22} & \textbf{CT} & \textbf{HL} & \textbf{TN} & \textbf{NN} & \textbf{FC} & \textbf{RT} & \textbf{PW} & \\
\midrule
\textbf{Human avg.} & 86.00 & 88.00 & 63.00 & 56.00 & 83.00 & 87.00  & 81.00  & 97.00 & 94.00 & 92.00 &84.00 & 82.00 & 82.75\\
\textbf{Human Ceiling} & 95.00 & 96.00 & 100.00 & 91.00 & 99.00 & 94.00  &  95.00  & 100.00 & 100.00 & 97.00 & 95.00 & 93.00 & 96.25 \\
\midrule
\textbf{RoBERTa} & 48.76 & 35.64  & 55.01 & 30.90 & 33.22  & 48.76  & 39.47  &  49.91   & \textbf{90.02}  & 32.01 & 53.50 & 55.92 & 47.76\\
\midrule
\textbf{LLaMA} & 19.31 & 26.35 & 17.81 & 17.98 & 18.41 & 24.10 & 32.26 & 41.91 & 47.29 & 40.00 & 48.89 & 53.78 & 32.34 \\
\textbf{Falcon} & 23.21 & 19.77 & 26.77 & 12.70 & 17.33 & 16.13 & 28.49 & 44.66 & 53.31 & 35.57 & 56.11 & 53.33 & 32.28 \\
\textbf{Mixtral-8x7B} & 45.29 & 36.81 & 48.92 & 41.40 & 38.97 & 50.84 & 33.27 & 40.86 & 50.13 & 32.08 & 46.84 & 44.80 & 42.52\\
\midrule
\textbf{ChatGPT} & 52.37 & 53.18 & 57.38 & 51.49  & 38.44  & 58.45  & 42.13  &  57.30  & 56.59 & 49.13 & {54.74} & 53.95 & 52.10 \\
\textbf{GPT-4} & 72.25 & 70.56 & \underline{87.20} & 73.12 & 58.49  & 56.40  & 46.01  & 60.08 & \underline{76.74} & \textbf{75.35} & 60.19  & 59.66 & 66.34 \\
\midrule
\textbf{o1 mini} & 69.35 & \underline{74.18} & 80.57 & 79.23 & 59.84 & \underline{75.12} & \textbf{63.69} & \underline{81.41} & 71.55 & \underline{63.07} & \underline{75.90} & 75.33 & 72.44 \\
\textbf{DeepSeek R1} & \underline{76.22} & \textbf{81.49} & 77.88 & \underline{90.01} & \underline{71.63} & \textbf{78.37} & \underline{62.05} & 75.74 & 72.58 & 59.96 & 75.29 & \underline{80.51} & \underline{75.14} \\
\textbf{QwQ-32B} & \textbf{85.70} & 73.33 & \textbf{93.76} & \textbf{92.35} & \textbf{86.30} & 73.87 & 61.53 & \textbf{81.96} & 75.59 & 62.45 & \textbf{78.10} & \textbf{82.40} & \textbf{78.95} \\
\bottomrule
\end{tabular}
}
\vspace{0.2em}
\caption{
LLMs' performance on the GLoRE benchmark. \emph{LQ}: LogiQA 2.0, \emph{RC}: ReClor, \emph{AL}: AR-LSAT, \emph{CT}: ConTRoL, \emph{HL}: HELP, \emph{TN}: TaxiNLI, \emph{NN}: NaN-NLI, \emph{FC}: FraCas, \emph{RT}: RuleTaker,  \emph{PW}: ProofWriter. All results are in \%, the best ones are in \textbf{bold}, and the second best ones are in \underline{underline}.
}
\label{Tab:result}
\vspace{-2em}
\end{table*}

\subsection{Main Results}

\noindent 
\textbf{Zero-shot results}.
Table \ref{Tab:result} summarizes the zero-shot evaluation results. The first block shows human performance.
The second block presents RoBERTa-base's supervised fine-tuning results. With 125M parameters, RoBERTa-base achieves 48.76\% and 33.22\% accuracy on LogiQA 2.0 and LogiQA22, respectively, lagging behind human performance. It performs better on NLI and TF tasks than MRC tasks, likely due to output ambiguities. On NaN-NLI, RoBERTa achieves 90.02\% accuracy, matching human performance, possibly due to learning superficial patterns from rule-based negation data. On ProofWriter, RoBERTa-base scores 55.92\%, indicating potential for specific logical reasoning tasks.

The third block shows zero-shot results for LLaMA, Falcon, and Mixtral. LLaMA and Falcon perform similarly (32.34\% vs. 32.28\%), suggesting comparable reasoning capabilities despite LLaMA-30B's smaller size. Both underperform RoBERTa-base on most tasks, except Falcon on RT. On MRC tasks, their accuracy is around 20\%, worse than random guessing in 4-way classification, indicating challenges in logical reasoning without in-context demonstrations. Performance gaps between LogiQA and LogiQA22 are smaller for these models, suggesting stable performance across data distributions without in-domain tuning. \textsc{Mixtral-8x7b} outperforms LLaMA and Falcon, demonstrating the effectiveness of mixture-of-expert models.

The fourth block provides zero-shot results for ChatGPT and GPT-4. Both models, especially GPT-4, surpass RoBERTa-base on several MRC benchmarks. However, GPT-4's accuracy drops significantly on LogiQA22 (58.49\% vs. 72.25\% on LogiQA 2.0), indicating sensitivity to data distribution. In NLI and TF tasks, ChatGPT and GPT-4 outperform RoBERTa, with ChatGPT achieving 58.45\% accuracy on ConTRoL, surpassing GPT-4. GPT-4's NLI performance varies across datasets, further highlighting its sensitivity to data distribution. TF task results show similar inconsistencies, suggesting model rationales differ from human reasoning.

The final block shows results for o1 mini, DeepSeek R1, and QwQ-32B, which achieve notable improvements over prior models. QwQ-32B attains the highest average accuracy (78.95\%), surpassing GPT-4 (66.34\%) and DeepSeek R1 (75.14\%). It achieves state-of-the-art results on MRC tasks like ReClor (93.76\%) and AR-LSAT (92.35\%), indicating the need for more challenging benchmarks for logical reasoning. Its robustness is evident in LogiQA22 (86.30\%), outperforming GPT-4 by 27.81 percentage points. However, QwQ-32B shows uneven performance on NLI datasets, such as HELP (61.53\%, lagging behind o1 mini’s 63.69\%), suggesting its reasoning capabilities are less generalizable in tasks requiring fine-grained entailment analysis.

While GPT-4 retains an advantage on FraCas (75.35\%), QwQ-32B surpasses GPT-4 on ReClor (93.76\% vs. 87.20\%), redefining state-of-the-art performance for MRC tasks. QwQ-32B and DeepSeek R1 showcase balanced performance across most tasks, with QwQ-32B achieving unprecedented TF results (e.g., 82.40\% on ProofWriter, outperforming both GPT-4’s 59.66\% and DeepSeek R1’s 80.51\%). Though still below the human average overall, these models mark substantial progress — QwQ-32B’s 78.95\% average accuracy (vs. DeepSeek R1’s 75.14\% and GPT-4’s 66.34\%) highlights significant architectural or training innovations for logical inference.

% 第七页表格和文字之间的空白太小了。[v]
\begin{table}[!t]
\small
\centering
\scalebox{1}{
\begin{tabular}{l|c|c|c|c}
\toprule
\textbf{Model} & \textbf{0-shot}  & \textbf{1-shot} & \textbf{2-shot} & {\bf 5-shot}  \\
\midrule
LLaMA  &  32.34 & 32.89 & 35.03 & 39.62  \\
Falcon  & 32.28 & 33.14 & 33.76 & 35.72  \\
ChatGPT  & 52.10 & 55.85 & 57.43 & 60.32  \\
GPT-4  &  66.34 & 70.31 & 71.44 & 75.83   \\

\bottomrule
\end{tabular}
}
\vspace{0.2em}
\caption{\label{Tab:few-shot}
Average accuracies on GLoRE few-shot evaluation. }
\vspace{-2em}
\end{table}

\textbf{Few-shot results}.
LLMs excel at in-context learning~\cite{dong2023survey}, where performance improves with context examples and demonstration methods~\cite{liu2021makes}. For this study, we randomly sampled instances (1 for 1-shot, 2 for 2-shot, and 5 for 5-shot) from each dataset and appended them to the prompt. We used the same model configuration as in the zero-shot scenario.
Table~\ref{Tab:few-shot} highlights the impact of in-context learning (ICL), as seen in GPT-4's 9\% accuracy gain with 5-shot learning. However, this improvement stems from statistical adaptation rather than true reasoning, as models rely on superficial patterns rather than human-like logical inference. This aligns with findings that chain-of-thought prompts correlate with outputs but do not causally drive reasoning \cite{bao2024likely}. While reasoning-enhanced models narrow the gap with human performance, their sensitivity to data distribution highlights the need for further research into more robust reasoning mechanisms. GLoRE's evolving framework will continue to track these advancements.

\subsection{Analysis}
% 我觉得4.3节下边的第一个自然段总括也许不应该放在这里。可以移动到第八页第二个自然段后边，作为第八页的第三个自然段，然后添加一个标题sensitivity to data distribution。[v]
% The above experiments show that the performance of LLMs is sensitive to the data distribution. Even though the underlying reasoning principles are the same, LLM performance varies significantly across datasets. This suggests that LLMs might not reason using the correct rationale but rely on superficial features. 

\noindent \textbf{Large language models vs. reasoning-enhanced models}.
The reasoning-enhanced models like QwQ-32B, DeepSeek R1, and OpenAI's o1 mini demonstrate significant improvements over traditional LLMs. QwQ-32B, in particular, achieves the highest average performance (78.95\%), indicating that its reinforcement learning framework or specialized training methodology enables better generalization across data distributions. 
While QwQ-32B dominates MRC and TF tasks, its relatively lower performance on NLI datasets like HELP (61.53\%) suggests that even state-of-the-art models struggle with tasks requiring monotonicity or negation reasoning, highlighting the need for broader evaluation beyond task-specific robustness.

\noindent \textbf{Data leakage concerns}.
While GLoRE includes diverse datasets, potential data leakage risks arise from overlapping sources. GPT-4’s lower accuracy on LogiQA22 (58.49\%) compared to LogiQA 2.0 (72.25\%) suggests limited exposure to newer data, reducing leakage concerns but highlighting distributional sensitivity.
The benchmark’s dynamic updates and inclusion of newly annotated datasets help mitigate leakage by testing models on unseen distributions.

\noindent \textbf{Sensitivity to data distribution}.
The above experiments show that the performance of LLMs is sensitive to the data distribution. Even though the underlying reasoning principles are the same, LLM performance varies significantly across datasets. This suggests that LLMs might not reason using the correct rationale but rely on superficial features. 
% 然后下边关于上下文学习的作用那个黑体标题可以去掉了，因为它也是关于数据分布的。 [v]
% \noindent 
% \textbf{The effect of in-context learning}. 
%Few-shot learning aims to educate models on the data distribution with as few instances as possible. The few-shot evaluation tests the efficiency of models to solve similar problems. Evaluation results are shown in Table \ref{Tab:few-shot}. With the increase of in-context examples, the accuracy of each tested model on the GLoRE benchmark increases. The models we tested all show in-context learning abilities on the logic reasoning benchmark. Among them, GPT-4 witnesses the highest performance gain with over a 9 percent accuracy boost on the 5-shot scenario compared to zero-shot. 
% dyr: 这里需要换行吗 [ ]
As shown in Table~\ref{Tab:result}, although GPT-4 achieves near-human performance on datasets like ReClor (87.20\%) and NaN-NLI (75.74\%), it lags significantly on others (e.g., HELP at 46.01\%). This inconsistency mirrors the behavior of reasoning-enhanced models like DeepSeek R1, revealing a critical divergence from human reasoning: once humans master a reasoning pattern, their performance generalizes robustly, whereas LLMs remain sensitive to data-specific features.

\section{Conclusion}

We constructed GLoRE, a dynamic and comprehensive benchmark tailored for assessing the logical reasoning capabilities of advanced language models, including GPT-4 and various strong open-source LLMs across multiple reasoning tasks. 
Our findings indicate that QwQ-32B, a reasoning-enhanced model, sets a new state-of-the-art on the GLoRE benchmark, significantly narrowing the gap to human performance. This underscores the potential of targeted architectural and training innovations for logical reasoning. GLoRE will be continually maintained to track advancements in this rapidly evolving domain.

\bibliographystyle{splncs04}
\bibliography{custom}

\end{document}